\documentclass[11pt]{article}
\usepackage{emnlp2016}
\usepackage{times}
\usepackage[hyphens]{url}
\usepackage{hyperref}
\usepackage{latexsym}
\usepackage{amsmath,amssymb}
\usepackage{multirow}
\usepackage{color}
\usepackage{graphicx}
\usepackage{bbm}
\usepackage{xspace}
\usepackage{wasysym}
\usepackage{latexsym}
\usepackage{graphicx}
\usepackage{algorithmic}
\usepackage{float}
\usepackage{mathtools}
\usepackage[titletoc,title]{appendix}
\usepackage{pgfplots}
\usepackage{placeins}
\usepackage{tabularx}
\usepackage{underscore}
\usepgfplotslibrary{groupplots}

\emnlpfinalcopy

\newcolumntype{L}[1]{>{\raggedright\arraybackslash}p{#1}}
\newcolumntype{C}[1]{>{\centering\arraybackslash}p{#1}}
\newcolumntype{R}[1]{>{\raggedleft\arraybackslash}p{#1}}\mathtoolsset{showonlyrefs}
\definecolor{mbgreen}{RGB}{34,189,64}

\newcommand{\CC}[1]{\multicolumn{1}{c|}{#1}}

\newcommand{\paragramsl}{\textsc{paragram-sl999}\xspace}
\newcommand{\paragramphrase}{\textsc{paragram-phrase}\xspace}

\newcommand{\charagramphrase}{\textsc{charagram-phrase}\xspace}
\newcommand{\charagram}{\textsc{charagram}\xspace}
\newcommand{\charagramshort}{\textsc{char}\xspace}
\newcommand{\len}{m}
\newcommand{\charCNN}{charCNN\xspace}
\newcommand{\charLSTM}{charLSTM\xspace}

\newcommand{\ngram}{$n$-gram\xspace}
\newcommand{\ngrams}{$n$-grams\xspace}
\newcommand{\vect}[1]{\mbox{\boldmath $#1$}}\newcommand{\charmat}{W}
\newcommand{\nonlin}{h}
\newcommand{\indic}[1]{\mathbb{I}\!\left[#1\right]}
\newcommand{\ngramvoc}{V}
\newcommand{\chardim}{d}

\newcommand\norm[1]{\left\lVert#1\right\rVert}
\DeclareMathOperator*{\argmax}{argmax}

\newcommand{\wsall}{WS353\xspace}

\newcommand{\simlex}{SL999\xspace}
\title{\charagram: Embedding Words and Sentences via Character $n$-grams}

\author{
John Wieting
\ \ \ \  Mohit Bansal
\ \ \ \  Kevin Gimpel
\ \ \ \  Karen Livescu
\\
Toyota Technological Institute at Chicago, Chicago, IL, 60637, USA\\
\tt{\{jwieting,mbansal,kgimpel,klivescu\}@ttic.edu}
}

\date{}

\begin{document}
\maketitle
\begin{abstract}

We present \charagram embeddings, a simple approach for learning character-based compositional models to embed textual sequences.  
A word or sentence is represented using a character \ngram count vector, followed by a single nonlinear transformation to yield a low-dimensional embedding. 
We use three tasks for evaluation: word similarity, sentence similarity, and part-of-speech tagging.  
We demonstrate that \charagram embeddings outperform more complex architectures based on character-level recurrent and convolutional neural networks, achieving new state-of-the-art performance on several similarity tasks.\footnote{Trained models and code 
are available at \url{http://ttic.uchicago.edu/~wieting}.
}

\end{abstract}

\section{Introduction}
Representing textual sequences such as words and sentences is a fundamental component of natural language understanding systems. 
Many functional architectures have been proposed to model compositionality in word sequences, ranging from simple averaging~\cite{Mitchell:Lapata:2010,iyyer-EtAl:2015:ACL-IJCNLP} to functions with rich recursive structure~\cite{SocherEtAl2011:PoolRAE,tai2015improved,bowman-acl2016}. Most work uses words as the smallest units in the compositional architecture, often using pretrained word embeddings or learning them specifically for the task of interest~\cite{tai2015improved,he-gimpel-lin:2015:EMNLP}.

Some prior work has found benefit from using character-based
compositional models that encode arbitrary character sequences into vectors.
Examples include recurrent neural networks (RNNs) and convolutional neural networks (CNNs) on character sequences, showing improvements for several NLP tasks~\cite{ling-EtAl:2015:EMNLP2,DBLP:journals/corr/KimJSR15,ballesteros-dyer-smith:2015:EMNLP,dossantos-guimaraes:2015:NEWS2015}.
By sharing subword information across words, character models have the potential to better represent rare words and morphological variants. 

Our approach,  \charagram, uses a much simpler functional architecture. 
We represent a character sequence by a vector containing counts of character \ngrams, inspired by \newcite{huang2013learning}. 
This vector is embedded into a low-dimensional space using a single nonlinear transformation. 
This can be interpreted as learning embeddings of character \ngrams, which are learned so as to produce effective sequence embeddings when a summation is performed over the character \ngrams in the sequence. 

We consider three evaluations: word similarity,  sentence similarity, and part-of-speech tagging. On multiple word similarity datasets, \charagram outperforms RNNs and CNNs, achieving state-of-the-art performance on SimLex-999~\cite{HillRK14}.  
When evaluated on a large suite of sentence-level semantic textual similarity tasks, \charagram embeddings again outperform the RNN and CNN architectures as well as the \paragramphrase embeddings of \newcite{wieting-16-full}.
We also consider English part-of-speech (POS) tagging using the bidirectional long short-term memory tagger of \newcite{ling-EtAl:2015:EMNLP2}. The three architectures reach similar performance, though \charagram converges fastest to high accuracy. 

We perform extensive analysis of our \charagram embeddings. We find large gains in performance on rare words, showing the empirical benefit of subword modeling. 
We also compare performance across different character \ngram vocabulary sizes, finding that the semantic tasks benefit far more from large vocabularies than the syntactic task. However, even for challenging semantic similarity tasks, we still see strong performance with only a few thousand character \ngrams.

Nearest neighbors show that \charagram embeddings simultaneously address differences due to spelling variation, morphology, and word choice. 
Inspection of embeddings of particular character \ngrams reveals etymological links; e.g., \emph{die} is close to \emph{mort}. We release our resources to the community in the hope that \charagram can provide a strong baseline for subword-aware text representation.

\section{Related Work}
\label{sec:relwork}
We first review work on using subword information in word embedding models. 
The simplest approaches append subword features to word embeddings, letting the model learn how to use the subword information for particular tasks. 
Some added knowledge-based morphological features to word representations~\cite{alexandrescu-kirchhoff:2006:HLT-NAACL06-Short,el2013morpheme}. 
Others learned embeddings jointly for subword units and words, defining simple compositional architectures (often based on addition) to create word embeddings from subword embeddings~\cite{lazaridou-EtAl:2013:ACL2013,botha2014compositional,qiu-EtAl:2014:Coling1,chenjoint2015}. 

A recent trend is to use richer functional architectures to convert character sequences into word embeddings. 
\newcite{luong-socher-manning:2013:CoNLL-2013} used recursive models to compose morphs into word embeddings, using unsupervised morphological analysis. 
\newcite{ling-EtAl:2015:EMNLP2} used a bidirectional long short-term memory (LSTM) RNN on characters to embed arbitrary word types, showing strong performance for language modeling and POS tagging.  \newcite{ballesteros-dyer-smith:2015:EMNLP} used this model to represent words for dependency parsing. Several have used character-level RNN architectures for machine translation, whether for representing source or target words~\cite{ling2015charnmt,luong2016}, or for generating entire translations character-by-character~\cite{chung2016character}.

\newcite{sutskever2011generating} and \newcite{graves2013generating} used character-level RNNs for language modeling. Others trained character-level RNN language models to provide features for NLP tasks, including tokenization and segmentation~\cite{chrupala2013text,D13-1146}, and text normalization~\cite{P14-2111}. 

CNNs with character $n$-gram filters have been used to embed arbitrary word types for several tasks, including language modeling~\cite{DBLP:journals/corr/KimJSR15}, part-of-speech tagging~\cite{santos2014learning}, named entity recognition~\cite{dossantos-guimaraes:2015:NEWS2015}, text classification~\cite{zhang-charcnn-2015}, and machine translation~\cite{costajussa2016charnmt}. 
Combinations of CNNs and RNNs on characters have also been explored~\cite{exploring2016limits}. 

Most closely-related to our approach is the DSSM (instantiated variously as ``deep semantic similarity model'' or ``deep structured semantic model'') developed by \newcite{huang2013learning}. 
For an information retrieval task, they represented words using feature vectors containing counts of character \ngrams.
\newcite{sperr-niehues-waibel:2013:CVSC} used a very similar technique to represent words in neural language models for machine translation. Our \charagram embeddings are based on this same idea. We show this strategy to be extremely effective when applied to both words and sentences, outperforming character LSTMs like those used by \newcite{ling-EtAl:2015:EMNLP2} and character CNNs like those from  \newcite{DBLP:journals/corr/KimJSR15}. 

\section{Models} \label{sec:models}
We now describe models that embed textual sequences using their characters, including our \charagram model and the baselines that we compare to.  
We denote a character-based textual sequence by $x=\langle
x_1,x_2,...,x_{\len}\rangle$, which includes space characters between
words as well as special start-of-sequence and end-of-sequence
characters. We use $x_i^j$ to denote the subsequence of characters
from position $i$ to position $j$ inclusive, i.e., $x_i^j=\langle x_i, x_{i+1}, ..., x_j\rangle$, and we define $x_i^i=x_i$. 

Our \charagram model embeds a character sequence $x$ 
by adding the vectors of its character \ngrams 
followed by an elementwise nonlinearity: 
\begin{equation}
g_{\charagramshort}(x) = \nonlin\!\left(\vect{b} + \sum_{i=1}^{\len+1} \sum_{j=1+i-k}^i \!\!\! \indic{x_j^i\in \ngramvoc} \charmat^{x_j^i}\right)\label{eq:charagram}
\end{equation}
\noindent where $\nonlin$ is a nonlinear function, 
$\vect{b}\in\mathbb{R}^{\chardim}$ is a bias vector, 
$k$ is the maximum length of any character \ngram, 
$\indic{p}$ is an indicator function that returns 1 if $p$ is true and 0 otherwise, 
$\ngramvoc$ is the set of character \ngrams included in the model, and 
$\charmat^{x_j^i}\in\mathbb{R}^{\chardim}$ is the vector for character \ngram $x_j^i$. 

The set $\ngramvoc$ is used to restrict the model to a predetermined set (vocabulary) of character \ngrams. Below, we compare several choices for defining this set. The number of parameters in the model is $\chardim + \chardim|\ngramvoc|$. 
This model is based on the letter \ngram hashing technique developed by \newcite{huang2013learning} for their DSSM approach. One can also view Eq.~\eqref{eq:charagram} (as they did) as first populating a vector of length $|\ngramvoc|$ with counts of character \ngrams followed by a nonlinear transformation. 

We compare the \charagram model to two other models. 
First we consider LSTM architectures~\cite{hochreiter1997long} over the character sequence $x$, using the version from \newcite{gers2003learning}. We use a forward LSTM over the characters in $x$, then take the final LSTM hidden vector as the representation of $x$. Below we refer to this model as ``\charLSTM.''

We also compare to convolutional neural network (CNN) architectures, which we refer to below as ``\charCNN.'' We use the architecture from \newcite{kim-14} with a single convolutional layer followed by an optional fully-connected layer. We use filters of varying lengths of character \ngrams, using two primary configurations of filter sets, one of which is identical to that used by \newcite{DBLP:journals/corr/KimJSR15}. Each filter operates over the entire sequence of character \ngrams in $x$ and we use max pooling for each filter. We tune over the choice of nonlinearity for both the convolutional filters and for the optional fully-connected layer. We give more details below about filter sets, \ngram lengths, and nonlinearities. 

We note that using character \ngram convolutional filters is similar to our use of character \ngrams in the \charagram model. The difference is that, in the \charagram model, the \ngram must match exactly for its vector to affect the representation, while in the CNN each filter will affect the representation of all sequences (depending on the nonlinearity being used). So the \charagram model is able to learn precise vectors for particular character \ngrams with specific meanings, while there is pressure for the CNN filters to capture multiple similar patterns that recur in the data. Our qualitative analysis shows the specificity of the learned character \ngram vectors learned by the \charagram model. 

\section{Experiments}  \label{sec:exp}
We perform three sets of experiments. The goal of the first two (Section~\ref{sec:all-sim-expts}) is to produce embeddings for textual sequences such that the embeddings for paraphrases have high cosine similarity. 
Our third evaluation (Section~\ref{sec:tag-expt}) is a classification task, and follows the setup of the English part-of-speech tagging experiment from \newcite{ling-EtAl:2015:EMNLP2}. 

\subsection{Word and Sentence Similarity}
\label{sec:all-sim-expts}
We compare the ability of our models to capture semantic similarity for both words and sentences. We train 
on noisy paraphrase pairs from the Paraphrase Database (PPDB; Ganitkevitch et al., 2013)\nocite{GanitkevitchDC13} with an $L_2$ regularized contrastive loss objective function, following the training procedure of \newcite{wieting2015ppdb-short} and \newcite{wieting-16-full}. Key details are provided here, but see Appendix~\ref{sec:appendix-a} for a fuller description.

\subsubsection{Datasets}

For word similarity, we focus on two of the most commonly used datasets for evaluating semantic similarity of word embeddings: WordSim-353 (\wsall)~\cite{finkelstein2001placing} and SimLex-999 (\simlex)~\cite{HillRK14}. We also evaluate our best model on the Stanford Rare Word Similarity Dataset \cite{luong-socher-manning:2013:CoNLL-2013}. 

For sentence similarity, we evaluate on a diverse set of 22 textual similarity datasets, including all datasets from every SemEval semantic textual similarity (STS) task from 2012 to 2015. We also evaluate on the SemEval 2015 Twitter task \cite{xu2015semeval} and the SemEval 2014 SICK Semantic Relatedness task \cite{marelli2014semeval}. 
Given two sentences, the aim of the STS tasks is to predict their similarity on a 0-5 scale, where 0 indicates the sentences are on different topics and 5 indicates that they are completely equivalent. 

Each STS task consists of 4-6 datasets covering a wide variety of domains, including newswire, tweets, glosses, machine translation outputs, web forums, news headlines, image and video captions, among others. Most submissions for these tasks use supervised models that are trained and tuned on provided training data or similar datasets from older tasks. 
Further details are provided in the official task descriptions~\cite{agirre2012semeval,diab2013eneko,agirre2014semeval,agirre2015semeval}.

\subsubsection{Preliminaries}

For training data, we use pairs from PPDB. For word similarity experiments, we train on word pairs and for sentence similarity, we train on phrase pairs. PPDB comes in different sizes (S, M, L, XL, XXL, and XXXL), where each larger size subsumes all smaller ones. The pairs in PPDB are sorted by a confidence measure and so the smaller sets contain higher precision paraphrases.

Before training the \charagram model, we need to populate $\ngramvoc$, the vocabulary of character \ngrams included in the model. 
We obtain these from the training data used for the final models in each setting, which is either the lexical or phrasal section of PPDB XXL. 
We tune over whether to include the full sets of character \ngrams in these datasets or only those that appear more than once. 

When extracting \ngrams, we include spaces and add an extra space before and after each word or phrase in the training and evaluation data to ensure that the beginning and end of each word is represented. We note that strong performance can be obtained using far fewer character \ngrams; we explore the effects of varying the number of \ngrams and the \ngram orders in Section~\ref{sec:ablation}.

We used Adam~\cite{kingma2014adam} with a learning rate of 0.001 to learn the parameters in the following experiments. 

\subsubsection{Word Embedding Experiments} \label{sec:word-expt}

\paragraph{Training and Tuning}
For hyperparameter tuning, we used one epoch on  
the lexical section of PPDB XXL, which consists of 770,007 word pairs.
We used either \wsall or \simlex for model selection (reported below). We then took the selected hyperparameters and trained for 50 epochs 
to ensure that all models had a chance to converge.

Full details of our tuning procedure are provided in Appendix~\ref{sec:appendix-b}. In short, we tuned all models thoroughly, tuning the activation functions for \charagram and \charCNN, as well as the regularization strength, 
mini-batch size, and sampling type for all models. For  \charCNN, we experimented with two filter sets: one uses 175 filters for each \ngram size $\in \{2, 3, 4\}$, and the other uses the set of filters from \newcite{DBLP:journals/corr/KimJSR15}, consisting of 25 filters of size 1, 50 of size 2, 75 of size 3, 100 of size 4, 125 of size 5, and 150 of size 6. We also experimented with using dropout \cite{srivastava2014dropout} on the inputs of the last layer of the \charCNN model in place of $L_2$ regularization, as well as removing the last feedforward layer. Neither of these variations significantly improved performance on our suite of tasks for word or sentence similarity. However, using more filters does improve performance, seemingly linearly with the square of the number of filters.

\paragraph{Architecture Comparison}

The results are shown in Table~\ref{table:wordsim}. The \charagram model outperforms both the \charLSTM and \charCNN models, and also outperforms recent strong results on \simlex.

We also found that the \charCNN and \charLSTM models take far more epochs to converge than the \charagram model. We noted this trend across experiments and explore it further in Section~\ref{sec:convergence}.

\begin{table}
\centering
\scriptsize
\begin{tabular} {| l | c || l | l |}
\hline
Model & Tuned on & \CC{\wsall} & \CC{\simlex}  \\\hline
\hline
\multirow{2}{*}{\charCNN} & \simlex & 26.31 & 30.64  \\ 
& \wsall & 33.19 & 16.73  \\ 
\hline
\multirow{2}{*}{\charLSTM} & \simlex & 48.27 & 54.54    \\ 
& \wsall & 51.43 & 48.83  \\ 
\hline
\multirow{2}{*}{\charagram} & \simlex & 53.87 & \bf 63.33  \\ 
& \wsall & \bf 58.35 & 60.00  \\ 
\hline
inter-annotator agreement & - &  75.6 & 78 \\
\hline
\end{tabular}
\caption{Word similarity results (Spearman's $\rho$ $\times$ 100) on \wsall and \simlex. The inter-annotator agreement is the average Spearman's $\rho$ between a single annotator with the average over all other annotators.
}
\label{table:wordsim}
\end{table}

\paragraph{Comparison to Prior Work}

\begin{table}[h]
\setlength{\tabcolsep}{4pt}
\scriptsize
\centering
\begin{tabular} {| l | l |} \hline
Model & \CC{\simlex} \\
\hline
\newcite{hill2014embedding} & 52 \\
\newcite{schwartz-reichart-rappoport:2015:Conll} & 56 \\
\newcite{faruqui2015non} & 58 \\
\newcite{wieting2015ppdb-short} & 66.7 \\
\charagram (large) & \bf 70.6 \\
\hline
\end{tabular}
\caption{\label{table:rare}
Spearman's $\rho \times 100$ on \simlex. \charagram (large) refers to the \charagram model described in Section~\ref{sec:ablation}. This model contains 173,881 character embeddings, more than the 100,283 in the \charagram model used to obtain the results in Table~\ref{table:wordsim}.
}
\end{table}

We found that performance of \charagram on word similarity tasks can be improved by using more character \ngrams. This is explored in Section~\ref{sec:ablation}. Our best result from these experiments was obtained with the largest model we considered, which contains 173,881 \ngram embeddings. When using \wsall for model selection and training for 25 epochs, this model achieves 70.6 on \simlex.
To our knowledge, this is the best result reported on \simlex in this setting; Table~\ref{table:rare} shows comparable recent results. Note that a higher \simlex number is reported in \cite{mrkvsic2016counter}, but the setting is not comparable to ours as they started with embeddings tuned on \simlex.

Lastly, we evaluated our model on the Stanford Rare Word Similarity Dataset~\cite{luong-socher-manning:2013:CoNLL-2013}, using \simlex for model selection. We obtained a Spearman's $\rho$ of 47.1, which outperforms the 41.8 result from \newcite{soricut2015unsupervised} and is competitive with the 47.8 reported in \newcite{pennington2014glove}, despite only using PPDB for training.

\subsubsection{Sentence Embedding Experiments} \label{sec:sent-expt}

\begin{table*}[t!]
\setlength{\tabcolsep}{4pt}
\scriptsize
\centering
\begin{tabular} { | l || c | c | c || C{1.6cm} | C{1.6cm} | C{1.6cm} | C{1.6cm} |} 
\hline
Dataset & 50\% & 75\% & Max  & \charCNN & \charLSTM & \paragramphrase & \charagramphrase \\
\hline
STS 2012 Average & 54.5 & 59.5 & 70.3 & 56.5 & 40.1 & 58.5 & \bf 66.1\\
STS 2013 Average & 45.3 & 51.4 & 65.3 & 47.7 & 30.7 & \bf 57.7 & 57.2\\
STS 2014 Average & 64.7 & 71.4 & 76.7 & 64.7 & 46.8 & 71.5 & \bf 74.7\\
STS 2015 Average & 70.2 & 75.8 & 80.2 & 66.0 & 45.5 & 75.7 & \bf 76.1\\
2014 SICK & 71.4 & 79.9 & 82.8 & 62.9 & 50.3 & \bf 72.0 & 70.0\\
2015 Twitter & 49.9 & 52.5 & 61.9 & 48.6 & 39.9 & 52.7 & \bf 53.6\\
\hline
\bf Average & 59.7 & 65.6 & 73.6 & 59.2 & 41.9 & 66.2 & \bf 68.7\\
\hline
\end{tabular}
\caption{\label{table:phrasesim1}
Results on SemEval textual similarity datasets (Pearson's $r \times 100$). The highest score in each row is in boldface (omitting the official task score columns). 
}
\end{table*}

\paragraph{Training and Tuning} 
We did initial training of our models using one pass through PPDB XL, which consists of 3,033,753 unique phrase pairs. Following \newcite{wieting-16-full}, we use the annotated phrase pairs developed by \newcite{PavlickEtAl-2015:ACL:PPDB2.0} as our validation set, using Spearman's $\rho$ to rank the models. We then take the highest performing models and train on the 9,123,575 unique phrase pairs in the phrasal section of PPDB XXL for 10 epochs.

For all experiments, we fix the mini-batch size to 100, the margin $\delta$ to 0.4, and use MAX sampling (see Appendix~\ref{sec:appendix-a}). For the \charagram model, $\ngramvoc$ contains all 122,610 character \ngrams ($n\in \{2,3,4\}$) in the PPDB XXL phrasal section. The other tuning settings are the same as in Section~\ref{sec:word-expt}. 

For another baseline, we train the \paragramphrase model of \newcite{wieting-16-full}, tuning its regularization strength
over $\{10^{-5},10^{-6},10^{-7},10^{-8}\}$. The \paragramphrase model simply uses word averaging as its composition function, but outperforms many more complex models. 

In this section, we refer to our model as \charagramphrase because the input is a character sequence containing multiple words rather than only a single word as in Section~\ref{sec:word-expt}. Since the vocabulary $\ngramvoc$ is defined by the training data sequences, the \charagramphrase model includes character \ngrams that span multiple words, permitting it to capture some aspects of word order and word co-occurrence, which the \paragramphrase model is unable to do. 

We encountered difficulties training the \charLSTM and \charCNN models for this task. We tried several strategies to improve their chance at convergence, including clipping gradients, increasing training data, and experimenting with different optimizers and learning rates. We found success by using the original (confidence-based) ordering of the PPDB phrase pairs for the initial epoch of learning, then shuffling them for subsequent epochs.  This is similar to curriculum learning~\cite{bengio2009curriculum}. The higher-confidence phrase pairs tend to be shorter and have many overlapping words, possibly making them easier to learn from. 

\paragraph{Results} 

An abbreviated version of the sentence similarity results is shown in Table~\ref{table:phrasesim1}; Appendix~\ref{sec:appendix-c} contains the full results. For comparison, we report performance for the median (50\%), third quartile (75\%), and top-performing (Max) systems from the shared tasks. We observe strong performance for the \charagramphrase model. It always does better than the \charCNN and \charLSTM models, and outperforms the \paragramphrase model on 15 of the 22 tasks. Furthermore, \charagramphrase matches or exceeds the top-performing task-tuned systems on 5 tasks, and is within 0.003 on 2 more. The \charLSTM and \charCNN models are significantly worse, with the \charCNN
being the better of the two and beating \paragramphrase on 4 of the tasks.

We emphasize that there are many other models that could be compared to, such as an LSTM over word embeddings. This and many other models were explored by \newcite{wieting-16-full}. Their \paragramphrase model, which simply learns word embeddings within an averaging composition function, was among their best-performing models. We used this model in our experiments as a strongly-performing representative of their results. 

Lastly, we note other recent work that considers a similar transfer learning setting. The FastSent model~\cite{hill2016learning} uses the 2014 STS task as part of its evaluation and reports an average Pearson's $r$ of 61.3, much lower than the 74.7 achieved by \charagramphrase on the same datasets. 

\subsection{POS Tagging Experiments} \label{sec:tag-expt}

\begin{table}[t]
\centering
\scriptsize
\begin{tabular} {| l | c |}
\hline
Model & Accuracy (\%) \\
\hline
\charCNN & 97.02 \\
\charLSTM & 96.90 \\ 
\charagram & 96.99 \\
\charagram (2-layer) & \bf 97.10 \\
\hline
\end{tabular}
\caption{Results on part-of-speech tagging.
}
\label{table:pos}
\end{table}

We now consider part-of-speech (POS) tagging, since it has been used as a testbed for evaluating architectures for character-level word representations. It also differs from semantic similarity, allowing us to evaluate our architectures on a syntactic task. We replicate the POS tagging experimental setup of \newcite{ling-EtAl:2015:EMNLP2}. Their model uses a bidirectional LSTM over character embeddings to represent words.  They then use the resulting word representations in another bidirectional LSTM that predicts the tag for each word.
We replace their character bidirectional LSTM with our three architectures: \charCNN, \charLSTM, and \charagram. 

We use the Wall Street Journal portion of the Penn Treebank, using Sections 1-18 for training, 19-21 for tuning, and 22-24 for testing. We set the dimensionality of the character embeddings to 50 and that of the (induced) word representations to 150. For optimization, we use stochastic gradient descent with a mini-batch size of 100 sentences. The learning rate and momentum are set to 0.2 and 0.95 respectively. We train the models for 50 epochs, again to ensure that all models have an opportunity to converge.

The other settings for our models are mostly the same as for the word and sentence experiments (Section~\ref{sec:all-sim-expts}). 
We again use character \ngrams with $n\in \{2,3,4\}$, tuning over whether to include all 54,893 in the training data or only those that occur more than once. 
However, there are two minor differences from the previous sections. First, we add a single binary feature to indicate if the token contains a capital letter. 
Second, our tuning considers rectified linear units as the activation function for the
\charagram and \charCNN architectures.\footnote{We did not consider ReLU for the similarity experiments because the final embeddings are used directly to compute cosine similarities, which led to poor performance when restricting the embeddings to be non-negative.} 

The results are shown in Table~\ref{table:pos}.  Performance is similar across models. We found that adding a second fully-connected 150 dimensional layer to the \charagram model improved results slightly.\footnote{We also tried adding a second (300 dimensional) layer for the word and sentence embedding models and found that it hurt performance.} 

\subsection{Convergence} \label{sec:convergence}

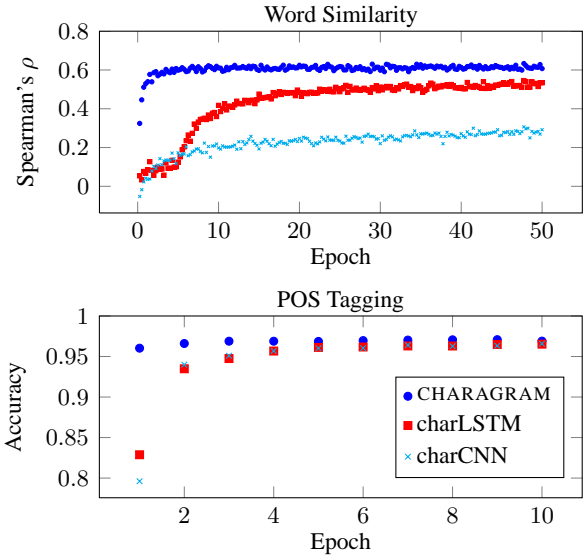
\begin{figure}[t]
\centering
\begin{tikzpicture}
\tikzstyle{every node}=[font=\footnotesize]
\begin{groupplot}[
    group style={
        group name=perf,
        group size=1 by 2,
        vertical sep=39pt,
    },
    xlabel = Epoch,
    height = 4cm,
    width = \columnwidth,
]
\nextgroupplot[title = Word Similarity, title style={at={(axis description cs:0.5,0.88)},anchor=south}, ymax=0.8, mark size = 0.8pt, ylabel = Spearman's $\rho$, ylabel near ticks, x label style={at={(axis description cs:0.5,0.08)},anchor=north}]
\addplot [color=blue,
    mark = *,
    only marks]
    table[x=x,y=y,col sep=comma] from {wsimchar.dat};
\addplot [color=red,
    mark = square*,
    only marks]
    table[x=x,y=y,col sep=comma] from {wsimlstm.dat};
\addplot [color=cyan,
    mark = x,
    only marks]
    table[x=x,y=y,col sep=comma] from {wsimcnn.dat};
\coordinate (top) at (rel axis cs:0,1);
\nextgroupplot[title = POS Tagging, title style={at={(axis description cs:0.5,0.88)},anchor=south}, ymax=1, mark size = 1.5pt, ylabel = Accuracy, ylabel near ticks, x label style={at={(axis description cs:0.5,0.08)},anchor=north}, legend style={at={(axis cs:10.5,0.8)},anchor=south east}, legend cell align=left]
\addplot [color=blue,
    mark = *,
    only marks]
    table[x=x,y=y,col sep=comma] from {poschar.dat};
    \addlegendentry{\charagram};
\addplot [color=red,
    mark = square*,
    only marks]
    table[x=x,y=y,col sep=comma] from {poslstm.dat};
    \addlegendentry{\charLSTM};
\addplot [color=cyan,
    mark = x,
    only marks]
    table[x=x,y=y,col sep=comma] from {poscnn.dat};
    \addlegendentry{\charCNN};
 \coordinate (bot) at (rel axis cs:1,0);
  \end{groupplot}
\path (top)--(bot) coordinate[midway] (group center);
\end{tikzpicture}
\caption{Plots of performance versus training epoch for word similarity and POS tagging.
}
\label{fig:plot}
\end{figure}

One observation we made during our experiments was that different models converged at significantly different rates. Figure~\ref{fig:plot} plots the performance of 
the word similarity and tagging tasks
as a function of the number of examples processed during training. For word similarity, we plot the oracle Spearman's $\rho$ on \simlex, while for tagging we plot tagging accuracy on the validation set. 
We evaluate performance every quarter epoch (approximately every 194,252 word pairs) for word similarity and every epoch for tagging. We only show the first 10 epochs of training in the tagging plot.

The plots show that the \charagram model converges quickly to high performance. The \charCNN and \charLSTM models take many more epochs to converge. Even with tagging, which uses a very high learning rate, \charagram converges significantly faster than the others. 
For word similarity, it appears that \charCNN and \charLSTM are still slowly improving at the end of 50 epochs. 
This suggests that if training was done for a much longer period, and possibly on more data, the \charLSTM or \charCNN models could match and surpass the \charagram model.
However, due to the large training sets available from PPDB and the computational requirements of these architectures, we were unable to explore the regime of training for many epochs. 
We conjecture that slow convergence could be the reason for the inferior performance of LSTMs for similarity tasks as reported by \newcite{wieting-16-full}. 

\subsection{Model Size Experiments} \label{sec:ablation}

The default setting for our \charagram and \charagramphrase models is to use all character bigram, trigrams, and 4-grams that occur in the training data at least $C$ times, tuning $C$ over the set $\{1, 2\}$. 
This results in a large number of parameters, which could be seen as an unfair advantage over the comparatively smaller \charCNN and \charLSTM models, which have up to 881,025 and 763,200 parameters respectively in the similarity experiments.\footnote{This includes 134 character embeddings.}

\begin{table}[t]
\centering
\scriptsize
\scalebox{0.8}{
    \begin{tabular}{|p{1.1cm}|R{1.17cm}|R{0.85cm}|R{0.85cm}|R{0.85cm}|R{1cm}|R{1cm}|} \hline
        Task & \multicolumn{1}{c|}{\# \ngrams} & \CC{2} & \CC{2,3} & \CC{2,3,4} & \CC{2,3,4,5} & \CC{2,3,4,5,6} \\ \hline
        \multirow{2}{*}{POS} & 100 & 95.52 & 96.09 & 96.15 & 96.13 & 96.16 \\
\multirow{2}{*}{Tagging} & 1,000 & 96.72 & 96.86 & 96.97 & 97.02 & 97.03 \\
        & 50,000 & 96.81 & 97.00 & 97.02 & 97.04 & \bf 97.09 \\\hline
\multirow{2}{*}{Word} & 100 & 6.2 & 7.0 & 7.7 & 9.1 & 8.8 \\
\multirow{2}{*}{Similarity} & 1,000 & 15.2 & 33.0 & 38.7 & 43.2 & 43.9 \\
        & 50,000 & 14.4 & 52.4 & 67.8 & 69.2 & \bf 69.5 \\\hline
\multirow{2}{*}{Sentence} & 100 & 40.2 & 33.8 & 32.5 & 31.2 & 29.8 \\
\multirow{2}{*}{Similarity} & 1,000 & 50.1 & 60.3 & 58.6 & 56.6 & 55.6 \\
        & 50,000 & 45.7 & 64.7 & \bf 66.6 & 63.0 & 61.3 \\\hline
    \end{tabular}
}
\caption{Results of using different numbers and different combinations of character \ngrams.
}
\label{table:ablation}
\end{table}

On the other hand, for a given training example, very few parameters in the \charagram model are actually used. For the \charCNN and \charLSTM models, by contrast, all parameters are used except the character embeddings for those characters that are not present in the example. 
For a sentence with 100 characters, and when using the 300-dimensional \charagram model with bigrams, trigrams, and 4-grams, there are approximately 90,000 parameters in use for this sentence,
far fewer than those used by the \charCNN and \charLSTM for the same sentence.

We performed a series of experiments to investigate how the \charagram and \charagramphrase models perform with different numbers and lengths of character \ngrams. For a given $k$, we took the top $k$ most frequent character \ngrams for each value of $n$ in use. We experimented with $k$ values in $\{100, 1000, 50000\}$. If there were fewer than $k$ unique character \ngrams for a given $n$, we used all of them. For these experiments, we did very little tuning, setting the regularization strength
to 0 and only tuning over the activation function. We repeated this experiment for all three of our tasks. For word similarity, we report performance on \simlex after training for 5 epochs on the lexical section of PPDB XXL. For sentence similarity, we report the average Pearson's $r$ over all 22 datasets after training for 5 epochs on the phrasal section of PPDB XL. For tagging, we report accuracy on the validation set after training for 50 epochs. The results are shown in Table~\ref{table:ablation}.

When using extremely small models with only 100 \ngrams of each order, we still see relatively strong performance on POS tagging.  However, the semantic similarity tasks require far more \ngrams to yield strong performance. Using 1000 \ngrams clearly outperforms 100, and 50,000 \ngrams performs best. 

\section{Analysis} \label{sec:analysis}
\subsection{Quantitative Analysis}

One of our primary motivations for character-based models is to address the issue of out-of-vocabulary (OOV) words, which were found to be one of the main sources of error for the \paragramphrase model from \newcite{wieting-16-full}. They reported a negative correlation (Pearson's $r$ of -0.45) between OOV rate and performance. We took the 12,108 sentence pairs in all 20 SemEval STS tasks and binned them by the total number of unknown words in the pairs.\footnote{Unknown words were defined as those not present in the 1.7 million unique (case-insensitive) tokens that comprise the vocabulary for the GloVe embeddings available at \url{http://nlp.stanford.edu/projects/glove/}. 
The \paragramsl embeddings, used to initialize the \paragramphrase model, use this same vocabulary.} 
We computed Pearson's $r$ over each bin. The results are shown in Table~\ref{table:oov}.

\begin{table}[h!]
\scriptsize
\centering
  \begin{tabular}{| C{1.95cm} | r | C{1.6cm} | C{1.7cm} |}
    \hline
    Number of Unknown Words & \multicolumn{1}{c|}{\multirow{2}{*}{$N$}} & \paragramphrase  & \charagramphrase\\
    \hline
0 & 11,292 & 71.4 & \bf 73.8\\
1 & 534 & 68.8 & \bf 78.8\\
2 & 194 & 66.4 & \bf 72.8\\
\hline
$ \geq 1$ & 816 & 68.6 & \bf 77.9\\
\hline
$ \geq 0$ & 12,108 & 71.0 & \bf 74.0\\
\hline
  \end{tabular}
  \caption{\label{table:oov}
Performance (Pearson's $r  \times 100$) as a function of the number of
unknown words in the sentence pairs over all 20 SemEval STS
datasets.  $N$ is the number of sentence pairs.
}
\end{table}

The \charagramphrase model has better performance for each number of unknown words. 
The \paragramphrase model degrades when more unknown words are present, presumably because it is forced to use the same unknown word embedding for all unknown words. The \charagramphrase model has no notion of unknown words, as it can embed any character sequence. 

We next investigated the sensitivity of the two models to length, as measured by the maximum of the lengths of the two sentences in a pair. We binned all of the 12,108 sentence pairs in the 20 SemEval STS tasks by length and then again found the Pearson's $r$ for both the \paragramphrase and \charagramphrase models. The results are shown in Table~\ref{table:length}. 

\begin{table}[h!]
\scriptsize
\centering
  \begin{tabular}{| c | r | C{1.8cm} | C{1.8cm} |}
    \hline
    \multirow{2}{*}{Max Length} &  \multicolumn{1}{c|}{\multirow{2}{*}{$N$}}  & \paragramphrase  & \charagramphrase\\
    \hline
$ \leq 4$ & 71 & 67.9 & \bf 72.9\\
5 & 216 & 71.1 & \bf 71.9\\
6 & 572 & 67.0 & \bf 69.7\\
7 & 1,097 & 71.5 & \bf 74.0\\
8 & 1,356 & 74.2 & \bf 74.5\\
9 & 1,266 & 71.7 & \bf 72.7\\
10 & 1,010 & 70.7 & \bf 74.2\\
11-15 & 3,143  & 71.8 & \bf 73.7\\
16-20 &  1,559 & 73.0 & \bf 75.1\\
$ \geq 21$ & 1,818 & 74.5 & \bf 75.4\\
 \hline
  \end{tabular}
  \caption{Performance (Pearson's $r  \times 100$) as a function of the maximum number of tokens in the sentence pairs over all 20 SemEval STS datasets. $N$ is the number of sentence pairs.
  }
\label{table:length}
\end{table}

We find that both models are robust to sentence length, achieving the highest correlations on the longest sentences. We also find that \charagramphrase outperforms \paragramphrase at all sentence lengths. 

\subsection{Qualitative Analysis}

\begin{table}[ht!]
	\centering
	\small
	\scalebox{0.7}{
	\begin{tabular}{|l|l|l|}
		\hline
		bigram &  \charagramphrase & \paragramphrase \\
		\hline
		not capable & incapable, unable, incapacity & not, capable, stalled \\
		not able & unable, incapable, incapacity & not, able, stalled \\
		not possible & impossible impracticable unable & not, stalled, possible \\
		not sufficient & insufficient, sufficient, inadequate & not, sufficient, stalled \\
		not easy & easy, difficult, tough & not, stalled, easy \\
		\hline
	\end{tabular}
	}
	\caption{\label{table:not} Nearest neighboring words of selected bigrams under \charagramphrase and \paragramphrase embeddings.
	}
\end{table}

\begin{table*}[th!]
	\centering
	\scriptsize
	\begin{tabular}{|l|l|}
		\hline
		Word &  \charagramphrase \\
		\hline
		vehicals & vehical, vehicles, vehicels, vehicular, cars, vehicle, automobiles, car \\
		serious-looking & serious, grave, acute, serious-minded, seriousness, gravity, serious-faced \\
		near-impossible & impossible, hard/impossible, audacious-impossible, impractical, unable \\
		growths & growth, grow, growing, increases, grows, increase, rise, growls, rising\\
		litered & liter, litering, lited, liters, literate, literature, literary, literal, lite, obliterated \\
		journeying & journey, journeys, voyage, trip, roadtrip, travel, tourney, voyages, road-trip \\
		babyyyyyy & babyyyyyyy, baby, babys, babe, baby.i, babydoll, babycake, darling \\
                 adirty & dirty, dirtyyyyyy, filthy, down-and-dirty, dirtying, dirties, ugly, dirty-blonde \\
		\hline
		\hline
		refunding & refunds, refunded, refund, repayment, reimbursement, rebate, repay\\
		 & reimbursements, reimburse, repaying, repayments, rebates, rebating, reimburses\\
		 \hline
                 professors & professor, professorships, professorship, teachers, professorial, teacher\\
                  & prof., teaches, lecturers, teachings, instructors, headteachers, teacher-student \\
                  \hline
                 huge & enormous, tremendous, large, big, vast, overwhelming, immense, giant\\
                  & formidable, considerable, massive, huger, large-scale, great, daunting \\
		\hline
	\end{tabular}
	\caption{\label{table:nn} Nearest neighbors of \charagramphrase embeddings. Above the double horizontal line are nearest neighbors of words that were not in our training data, and below it are nearest neighbors of words that were in our training data.
	}
\end{table*}

Aside from OOVs,
the \paragramphrase model 
lacks the ability to model word order or cooccurrence, 
since it simply averages the words in the sequence. We were interested to see whether \charagramphrase could handle negation, since it does model limited information about word order (via character \ngrams that span multiple words in the sequence). We made a list of ``not'' bigrams that could be represented by a single word, then embedded each bigram using both models and did a nearest-neighbor search over a working vocabulary.\footnote{This contained all words in PPDB-XXL, our evaluations, and in two other datasets: 
the Stanford Sentiment task~\cite{socher-13} and the SNLI dataset~\cite{bowman2015large}, resulting in 93,217 unique (up-to-casing) tokens.} The results, in Table~\ref{table:not}, show how the \charagramphrase embeddings model negation. In all cases but one, the nearest neighbor is a paraphrase for the 
bigram and the next neighbors are mostly paraphrases as well. The \paragramphrase model, unsurprisingly, is incapable of modeling negation. In all cases, the nearest neighbor is {\it not}, as this word carries much more weight than the word it modifies. The remaining nearest neighbors are either the modified word or {\it stalled}. 

We did two additional nearest neighbor explorations with our \charagramphrase model. 
In the first, we collected the nearest neighbors for words that were not in the training data (i.e. PPDB XXL), but were in our working vocabulary. This consisted of 59,660 words. In the second, we collected nearest neighbors of words that were in our training data which consisted of 37,765 tokens. 

A sample of the nearest neighbors is shown in Table~\ref{table:nn}. Several kinds of similarity are being captured simultaneously by the model. 
One kind is similarity in terms of spelling variation, including misspellings ({\it vehicals}, {\it vehicels}, and {\it vehicles}) and repetition for emphasis ({\it baby} and {\it babyyyyyyy}). Another kind is similarity in terms of morphological variants of a shared root (e.g., {\it journeying} and {\it journey}). We also see that the model has learned many strong synonym relationships without significant amounts of overlapping \ngrams (e.g., {\it vehicles}, {\it cars}, and {\it automobiles}). We find these characteristics for words both in and out of the training data. 
Words in the training data, which tend to be more commonly used, do tend to have higher precision in their nearest neighbors (e.g., see neighbors for {\it huge}). We noted occasional mistakes for words that share a large number of \ngrams but are not paraphrases (see nearest neighbors for {\it litered} which is likely a misspelling of {\it littered}).

\begin{table}[h!]
	\centering
	\scriptsize
	\begin{tabular}{|l|l|l|}
		\hline
		\ngram & \ngram Embedding \\
		\hline
		die & _dy, _die, dead,_dyi, rlif, mort, ecea, rpse, d_aw \\
		foo &_foo,_eat, meal, alim, trit, feed, grai,_din, nutr, toe \\
		pee & peed, hast, spee, fast, mpo_, pace, _vel, loci, ccel \\
		aiv & waiv, aive, boli, epea, ncel, abol, lift, bort, bol \\
		ngu & ngue, uist, ongu, tong, abic, gual, fren, ocab, ingu \\
 		_2 & 2_, _02, _02_,_tw,_dua,_xx,_ii_, xx, o_14, d_.2\\
		\hline
	\end{tabular}
	\caption{\label{table:noncol} Nearest neighbors of character \ngram embeddings from our trained \charagramphrase model. The underscore indicates a space, which signals the beginning or end of a word.
	}
\end{table}

Lastly, since our model learns embeddings for character \ngrams, 
we include an analysis of character \ngram nearest neighbors in Table~\ref{table:noncol}. These \ngrams appear to be grouped into themes, such as death (first row), food (second row), and speed (third row), 
but have different granularities. The \ngrams in the last row appear in paraphrases of {\it 2}, whereas the second-to-last row shows \ngrams in words like {\it french} and {\it vocabulary}, which can broadly be classified as having to do with language.

\section{Conclusion}

We performed a careful empirical comparison of character-based compositional architectures on three NLP tasks. 
While most prior work has considered machine translation, language modeling, and syntactic analysis, we showed how character-level modeling can improve semantic similarity tasks, both quantitatively and with extensive qualitative analysis. 
We found a consistent trend: the simplest architecture converges fastest to high performance. These results, coupled with those from \newcite{wieting-16-full}, suggest that practitioners should begin with simple architectures rather than moving immediately to RNNs and CNNs.  We release our code and trained models so they can be used by the NLP community for general-purpose, character-based text representation. 

\section*{Acknowledgments}

We would like to thank the developers of Theano~\cite{2016arXiv160502688short} and NVIDIA Corporation for donating GPUs used in this research.

\begin{appendices}
\section{Training} \label{sec:appendix-a}
\begin{table*}
\setlength{\tabcolsep}{4pt}
\scriptsize
\centering
\begin{tabular} { | l || c | c | c || C{1.6cm} | C{1.6cm} | C{1.6cm} | C{1.6cm} |} 
\hline
Dataset & 50\% & 75\% & Max  & \charCNN & \charLSTM & \paragramphrase & \charagramphrase \\
\hline
MSRpar & 51.5 & 57.6 & 73.4 & 50.6 & 23.6 & 42.9 & \bf 59.7\\
MSRvid & 75.5 & 80.3 & 88.0 & 72.2 & 47.2 & 76.1 & \bf 79.6\\
SMT-eur & 44.4 & 48.1 & 56.7 & 50.9 & 38.5 & 45.5 & \bf 57.2\\
OnWN & 60.8 & 65.9 & 72.7 & 61.8 & 53.0 & \bf 70.7 & 68.7\\
SMT-news & 40.1 & 45.4 & 60.9 & 46.8 & 38.3 & 57.2 & \bf 65.2\\
\hline
STS 2012 Average & 54.5 & 59.5 & 70.3 & 56.5 & 40.1 & 58.5 & \bf 66.1\\
\hline
headline & 64.0 & 68.3 & 78.4 & 68.1 & 54.4 & 72.3 & \bf 75.0\\
OnWN & 52.8 & 64.8 & 84.3 & 54.4 & 33.5 & \bf 70.5 & 67.8\\
FNWN & 32.7 & 38.1 & 58.2 & 26.4 & 10.6 & \bf 47.5 & 42.3\\
SMT & 31.8 & 34.6 & 40.4 & 42.0 & 24.2 & 40.3 & \bf 43.6\\
\hline
STS 2013 Average & 45.3 & 51.4 & 65.3 & 47.7 & 30.7 & \bf 57.7 & 57.2\\
\hline
deft forum & 36.6 & 46.8 & 53.1 & 45.6 & 19.4 & 50.2 & \bf 62.7\\
deft news & 66.2 & 74.0 & 78.5 & 73.5 & 54.6 & 73.2 & \bf 77.0\\
headline & 67.1 & 75.4 & 78.4 & 67.4 & 53.7 & 69.1 & \bf 74.3\\
images & 75.6 & 79.0 & 83.4 & 68.7 & 53.6 & \bf 80.0 & 77.6\\
OnWN & 78.0 & 81.1 & 87.5 & 66.8 & 46.1 & \bf 79.9 & 77.0\\
tweet news & 64.7 & 72.2 & 79.2 & 66.2 & 53.6 & 76.8 & \bf 79.1\\
\hline
STS 2014 Average & 64.7 & 71.4 & 76.7 & 64.7 & 46.8 & 71.5 & \bf 74.7\\
\hline
answers-forums & 61.3 & 68.2 & 73.9 & 47.2 & 27.3 & \bf 67.4 & 61.5\\
answers-students & 67.6 & 73.6 & 78.8 & 75.0 & 63.1 & 78.3 & \bf 78.5\\
belief & 67.7 & 72.2 & 77.2 & 65.7 & 22.6 & 76.0 & \bf 77.2\\
headline & 74.2 & 80.8 & 84.2 & 72.2 & 61.7 & 74.5 & \bf 78.7\\
images & 80.4 & 84.3 & 87.1 & 70.0 & 52.8 & 82.2 & \bf 84.4\\
\hline
STS 2015 Average & 70.2 & 75.8 & 80.2 & 66.0 & 45.5 & 75.7 & \bf 76.1\\
\hline
2014 SICK & 71.4 & 79.9 & 82.8 & 62.9 & 50.3 & \bf 72.0 & 70.0\\
\hline
2015 Twitter & 49.9 & 52.5 & 61.9 & 48.6 & 39.9 & 52.7 & \bf 53.6\\
\hline
\bf Average & 59.7 & 65.6 & 73.6 & 59.2 & 41.9 & 66.2 & \bf 68.7\\
\hline
\end{tabular}
\caption{\label{table:phrasesim2}
Results on SemEval textual similarity datasets (Pearson's $r \times 100$). The highest score in each row is in boldface (omitting the official task score columns).
}
\end{table*}

For word and sentence similarity, we follow the training procedure of \newcite{wieting2015ppdb-short} and \newcite{wieting-16-full}, described below. For part-of-speech tagging, we follow the English Penn Treebank training procedure of \newcite{ling-EtAl:2015:EMNLP2}. 

For the similarity tasks, the training data consists of a set $X$ of phrase pairs $\langle x_1, x_2\rangle$ from the Paraphrase Database (PPDB; Ganitkevitch et al., 2013),\nocite{GanitkevitchDC13} where $x_1$ and $x_2$ are assumed to be paraphrases. 
We optimize a margin-based loss: 

\begin{small}
\begin{align}
&\underset{\theta}{\text{min}} \frac{1}{|X|}\Bigg(\sum_{\langle x_1,x_2\rangle \in X} 
\max(0,\delta - \cos(g(x_1), g(x_2))\\
&+ \cos(g(x_1), g(t_1))) + \max(0,\delta - \cos(g(x_1),g(x_2))\\
&+ \cos(g(x_2), g(t_2)))\bigg) + \lambda \norm{\theta}^2 
\label{eq:phrase}
\end{align} 
\end{small}

\noindent where $g$ is the embedding function in use, $\delta$ is the margin, the full set of parameters is contained in $\theta$ (e.g., for the \charagram model, $\theta = \langle \charmat, \vect{b}\rangle$), $\lambda$ 
is the $L_2$  regularization coefficient, and $t_1$ and $t_2$ are carefully selected negative examples taken from a mini-batch during optimization (discussed below). Intuitively, we want the two phrases to be more similar to each other ($\cos(g(x_1), g(x_2))$) than either is to their respective negative examples $t_1$ and $t_2$, by a margin of at least $\delta$. 

\subsection{Selecting Negative Examples}
To select $t_1$ and $t_2$ in Eq. 2, we tune the choice between two approaches. The first, MAX, simply chooses the most similar phrase in some set of phrases (other than those in the given phrase pair). For simplicity and to reduce the number
of tunable parameters, we use the mini-batch for this set, but it could be a separate set.  Formally, MAX corresponds to choosing $t_1$ for a given $\langle x_1, x_2\rangle$ as follows:
\begin{equation}
t_1 = \argmax_{t : \langle t, \cdot\rangle \in X_b \setminus \{\langle x_1, x_2\rangle\}} \cos(g(x_1), g(t))
\end{equation}
\noindent where $X_b\subseteq X$ is the current mini-batch. 
That is, we want to choose a negative example $t_i$ that is similar to $x_i$ according to the current model parameters. 
The downside of this approach is that we may occasionally choose a phrase $t_i$ that is actually a true paraphrase of $x_i$. 

The second strategy selects negative examples using MAX with probability 0.5 and selects them randomly from the mini-batch otherwise. We call this sampling strategy MIX. We tune over the choice of strategy in our experiments. 
\section{Tuning Word Similarity Models} \label{sec:appendix-b}
For all architectures, we tuned over the mini-batch size (25 or 50) and the type of sampling used (MIX or MAX). $\delta$ was set to 0.4 and the dimensionality $\chardim$ of each model was set to 300.

For the \charagram model, we tuned the activation function $\nonlin$ ($\tanh$ or linear) and regularization coefficient $\lambda$ (over $\{10^{-4},10^{-5},10^{-6}\}$). The \ngram vocabulary $\ngramvoc$ contained all 100,283 character \ngrams ($n\in \{2,3,4\}$) in the lexical section of PPDB XXL. 

For \charCNN and \charLSTM, we randomly initialized 300 dimensional character embeddings for all unique characters in the training data. For \charLSTM, we tuned over whether to include an output gate. For  \charCNN, we tuned the filter activation function (rectified linear or $\tanh$) and tuned the activation for the fully-connected layer ($\tanh$ or linear).
For both the \charLSTM and \charCNN models, we tuned $\lambda$ over $\{10^{-4},10^{-5},10^{-6}\}$.
\section{Full Sentence Similarity Results} \label{sec:appendix-c}
Table~\ref{table:phrasesim2} shows the full results of our sentence similarity experiments.
\end{appendices}

\bibliography{emnlp2016}

\begin{thebibliography}{}

\bibitem[\protect\citename{Agirre \bgroup et al.\egroup
  }2012]{agirre2012semeval}
Eneko Agirre, Mona Diab, Daniel Cer, and Aitor Gonzalez-Agirre.
\newblock 2012.
\newblock {SemEval}-2012 task 6: A pilot on semantic textual similarity.
\newblock In {\em Proceedings of the First Joint Conference on Lexical and
  Computational Semantics-Volume 1: Proceedings of the main conference and the
  shared task, and Volume 2: Proceedings of the Sixth International Workshop on
  Semantic Evaluation}. Association for Computational Linguistics.

\bibitem[\protect\citename{Agirre \bgroup et al.\egroup }2013]{diab2013eneko}
Eneko Agirre, Daniel Cer, Mona Diab, Aitor Gonzalez-Agirre, and Weiwei Guo.
\newblock 2013.
\newblock *{SEM} 2013 shared task: Semantic textual similarity.
\newblock In {\em Second Joint Conference on Lexical and Computational
  Semantics (*{SEM}), Volume 1: Proceedings of the Main Conference and the
  Shared Task: Semantic Textual Similarity}.

\bibitem[\protect\citename{Agirre \bgroup et al.\egroup
  }2014]{agirre2014semeval}
Eneko Agirre, Carmen Banea, Claire Cardie, Daniel Cer, Mona Diab, Aitor
  Gonzalez-Agirre, Weiwei Guo, Rada Mihalcea, German Rigau, and Janyce Wiebe.
\newblock 2014.
\newblock {SemEval}-2014 task 10: Multilingual semantic textual similarity.
\newblock In {\em Proceedings of the 8th International Workshop on Semantic
  Evaluation ({SemEval} 2014)}.

\bibitem[\protect\citename{Agirre \bgroup et al.\egroup
  }2015]{agirre2015semeval}
Eneko Agirre, Carmen Banea, Claire Cardie, Daniel Cer, Mona Diab, Aitor
  Gonzalez-Agirre, Weiwei Guo, Inigo Lopez-Gazpio, Montse Maritxalar, Rada
  Mihalcea, German Rigau, Larraitz Uria, and Janyce Wiebe.
\newblock 2015.
\newblock {SemEval}-2015 task 2: Semantic textual similarity, {English},
  {Spanish} and pilot on interpretability.
\newblock In {\em Proceedings of the 9th International Workshop on Semantic
  Evaluation ({SemEval} 2015)}.

\bibitem[\protect\citename{Alexandrescu and
  Kirchhoff}2006]{alexandrescu-kirchhoff:2006:HLT-NAACL06-Short}
Andrei Alexandrescu and Katrin Kirchhoff.
\newblock 2006.
\newblock Factored neural language models.
\newblock In {\em Proceedings of the Human Language Technology Conference of
  the NAACL, Companion Volume: Short Papers}, pages 1--4, New York City, USA,
  June. Association for Computational Linguistics.

\bibitem[\protect\citename{Ballesteros \bgroup et al.\egroup
  }2015]{ballesteros-dyer-smith:2015:EMNLP}
Miguel Ballesteros, Chris Dyer, and Noah~A. Smith.
\newblock 2015.
\newblock Improved transition-based parsing by modeling characters instead of
  words with lstms.
\newblock In {\em Proceedings of the 2015 Conference on Empirical Methods in
  Natural Language Processing}, pages 349--359, Lisbon, Portugal, September.
  Association for Computational Linguistics.

\bibitem[\protect\citename{Bengio \bgroup et al.\egroup
  }2009]{bengio2009curriculum}
Yoshua Bengio, J{\'e}r{\^o}me Louradour, Ronan Collobert, and Jason Weston.
\newblock 2009.
\newblock Curriculum learning.
\newblock In {\em Proceedings of the 26th annual international conference on
  machine learning}, pages 41--48. ACM.

\bibitem[\protect\citename{Botha and Blunsom}2014]{botha2014compositional}
Jan~A. Botha and Phil Blunsom.
\newblock 2014.
\newblock Compositional morphology for word representations and language
  modelling.
\newblock In {\em International Conference on Machine Learning (ICML)}.

\bibitem[\protect\citename{Bowman \bgroup et al.\egroup }2015]{bowman2015large}
Samuel~R Bowman, Gabor Angeli, Christopher Potts, and Christopher~D Manning.
\newblock 2015.
\newblock A large annotated corpus for learning natural language inference.
\newblock {\em arXiv preprint arXiv:1508.05326}.

\bibitem[\protect\citename{Bowman \bgroup et al.\egroup }2016]{bowman-acl2016}
Samuel~R. Bowman, Jon Gauthier, Abhinav Rastogi, Raghav Gupta, Christopher~D.
  Manning, and Christopher Potts.
\newblock 2016.
\newblock A fast unified model for parsing and sentence understanding.
\newblock In {\em Proceedings of ACL}.

\bibitem[\protect\citename{Chen \bgroup et al.\egroup }2015]{chenjoint2015}
Xinxiong Chen, Lei Xu, Zhiyuan Liu, Maosong Sun, and Huanbo Luan.
\newblock 2015.
\newblock Joint learning of character and word embeddings.
\newblock In {\em Proceedings of International Joint Conference on Artificial
  Intelligence (IJCAI)}.

\bibitem[\protect\citename{Chrupa{\l}a}2013]{chrupala2013text}
Grzegorz Chrupa{\l}a.
\newblock 2013.
\newblock Text segmentation with character-level text embeddings.
\newblock {\em arXiv preprint arXiv:1309.4628}.

\bibitem[\protect\citename{Chrupa{\l}a}2014]{P14-2111}
Grzegorz Chrupa{\l}a.
\newblock 2014.
\newblock Normalizing tweets with edit scripts and recurrent neural embeddings.
\newblock In {\em Proceedings of the 52nd Annual Meeting of the Association for
  Computational Linguistics (Volume 2: Short Papers)}, pages 680--686.
  Association for Computational Linguistics.

\bibitem[\protect\citename{Chung \bgroup et al.\egroup
  }2016]{chung2016character}
Junyoung Chung, Kyunghyun Cho, and Yoshua Bengio.
\newblock 2016.
\newblock A character-level decoder without explicit segmentation for neural
  machine translation.
\newblock {\em arXiv preprint arXiv:1603.06147}.

\bibitem[\protect\citename{Costa{-}Juss{\`{a}} and
  Fonollosa}2016]{costajussa2016charnmt}
Marta~R. Costa{-}Juss{\`{a}} and Jos{\'{e}} A.~R. Fonollosa.
\newblock 2016.
\newblock Character-based neural machine translation.
\newblock {\em arXiv preprint arXiv:1603.00810}.

\bibitem[\protect\citename{dos Santos and
  Guimar\~{a}es}2015]{dossantos-guimaraes:2015:NEWS2015}
Cicero dos Santos and Victor Guimar\~{a}es.
\newblock 2015.
\newblock Boosting named entity recognition with neural character embeddings.
\newblock In {\em Proceedings of the Fifth Named Entity Workshop}, pages
  25--33, Beijing, China, July. Association for Computational Linguistics.

\bibitem[\protect\citename{dos Santos and Zadrozny}2014]{santos2014learning}
Cicero~N. dos Santos and Bianca Zadrozny.
\newblock 2014.
\newblock Learning character-level representations for part-of-speech tagging.
\newblock In {\em Proceedings of the 31st International Conference on Machine
  Learning (ICML-14)}, pages 1818--1826.

\bibitem[\protect\citename{El-Desoky~Mousa \bgroup et al.\egroup
  }2013]{el2013morpheme}
Amr El-Desoky~Mousa, Hong-Kwang~Jeff Kuo, Lidia Mangu, and Hagen Soltau.
\newblock 2013.
\newblock Morpheme-based feature-rich language models using deep neural
  networks for lvcsr of egyptian arabic.
\newblock In {\em 2013 IEEE International Conference on Acoustics, Speech and
  Signal Processing (ICASSP)}, pages 8435--8439. IEEE.

\bibitem[\protect\citename{Evang \bgroup et al.\egroup }2013]{D13-1146}
Kilian Evang, Valerio Basile, Grzegorz Chrupa{\l}a, and Johan Bos.
\newblock 2013.
\newblock Elephant: Sequence labeling for word and sentence segmentation.
\newblock In {\em Proceedings of the 2013 Conference on Empirical Methods in
  Natural Language Processing}, pages 1422--1426. Association for Computational
  Linguistics.

\bibitem[\protect\citename{Faruqui and Dyer}2015]{faruqui2015non}
Manaal Faruqui and Chris Dyer.
\newblock 2015.
\newblock Non-distributional word vector representations.
\newblock {\em arXiv preprint arXiv:1506.05230}.

\bibitem[\protect\citename{Finkelstein \bgroup et al.\egroup
  }2001]{finkelstein2001placing}
Lev Finkelstein, Evgeniy Gabrilovich, Yossi Matias, Ehud Rivlin, Zach Solan,
  Gadi Wolfman, and Eytan Ruppin.
\newblock 2001.
\newblock Placing search in context: The concept revisited.
\newblock In {\em Proceedings of the 10th international conference on World
  Wide Web}. ACM.

\bibitem[\protect\citename{Ganitkevitch \bgroup et al.\egroup
  }2013]{GanitkevitchDC13}
Juri Ganitkevitch, Benjamin~Van Durme, and Chris Callison-Burch.
\newblock 2013.
\newblock Ppdb: The paraphrase database.
\newblock In {\em HLT-NAACL}. The Association for Computational Linguistics.

\bibitem[\protect\citename{Gers \bgroup et al.\egroup }2003]{gers2003learning}
Felix~A Gers, Nicol~N Schraudolph, and J{\"u}rgen Schmidhuber.
\newblock 2003.
\newblock Learning precise timing with lstm recurrent networks.
\newblock {\em The Journal of Machine Learning Research}, 3.

\bibitem[\protect\citename{Graves}2013]{graves2013generating}
Alex Graves.
\newblock 2013.
\newblock Generating sequences with recurrent neural networks.
\newblock {\em arXiv preprint arXiv:1308.0850}.

\bibitem[\protect\citename{He \bgroup et al.\egroup
  }2015]{he-gimpel-lin:2015:EMNLP}
Hua He, Kevin Gimpel, and Jimmy Lin.
\newblock 2015.
\newblock Multi-perspective sentence similarity modeling with convolutional
  neural networks.
\newblock In {\em Proceedings of the 2015 Conference on Empirical Methods in
  Natural Language Processing}.

\bibitem[\protect\citename{Hill \bgroup et al.\egroup }2014]{hill2014embedding}
Felix Hill, Kyunghyun Cho, Sebastien Jean, Coline Devin, and Yoshua Bengio.
\newblock 2014.
\newblock Embedding word similarity with neural machine translation.
\newblock {\em arXiv preprint arXiv:1412.6448}.

\bibitem[\protect\citename{Hill \bgroup et al.\egroup }2015]{HillRK14}
Felix Hill, Roi Reichart, and Anna Korhonen.
\newblock 2015.
\newblock {SimLex}-999: Evaluating semantic models with (genuine) similarity
  estimation.
\newblock {\em Computational Linguistics}, 41(4).

\bibitem[\protect\citename{Hill \bgroup et al.\egroup }2016]{hill2016learning}
Felix Hill, Kyunghyun Cho, and Anna Korhonen.
\newblock 2016.
\newblock Learning distributed representations of sentences from unlabelled
  data.
\newblock {\em arXiv preprint arXiv:1602.03483}.

\bibitem[\protect\citename{Hochreiter and Schmidhuber}1997]{hochreiter1997long}
Sepp Hochreiter and J{\"u}rgen Schmidhuber.
\newblock 1997.
\newblock Long short-term memory.
\newblock {\em Neural computation}, 9(8).

\bibitem[\protect\citename{Huang \bgroup et al.\egroup
  }2013]{huang2013learning}
Po-Sen Huang, Xiaodong He, Jianfeng Gao, Li~Deng, Alex Acero, and Larry Heck.
\newblock 2013.
\newblock Learning deep structured semantic models for web search using
  clickthrough data.
\newblock In {\em Proceedings of the 22nd ACM international conference on
  Conference on information \& knowledge management}, pages 2333--2338. ACM.

\bibitem[\protect\citename{Iyyer \bgroup et al.\egroup
  }2015]{iyyer-EtAl:2015:ACL-IJCNLP}
Mohit Iyyer, Varun Manjunatha, Jordan Boyd-Graber, and Hal Daum\'{e}~III.
\newblock 2015.
\newblock Deep unordered composition rivals syntactic methods for text
  classification.
\newblock In {\em Proceedings of the 53rd Annual Meeting of the Association for
  Computational Linguistics and the 7th International Joint Conference on
  Natural Language Processing (Volume 1: Long Papers)}.

\bibitem[\protect\citename{J{\'{o}}zefowicz \bgroup et al.\egroup
  }2016]{exploring2016limits}
Rafal J{\'{o}}zefowicz, Oriol Vinyals, Mike Schuster, Noam Shazeer, and Yonghui
  Wu.
\newblock 2016.
\newblock Exploring the limits of language modeling.
\newblock {\em CoRR}, abs/1602.02410.

\bibitem[\protect\citename{Kim \bgroup et al.\egroup
  }2015]{DBLP:journals/corr/KimJSR15}
Yoon Kim, Yacine Jernite, David Sontag, and Alexander~M. Rush.
\newblock 2015.
\newblock Character-aware neural language models.
\newblock {\em CoRR}, abs/1508.06615.

\bibitem[\protect\citename{Kim}2014]{kim-14}
Yoon Kim.
\newblock 2014.
\newblock Convolutional neural networks for sentence classification.
\newblock In {\em Proceedings of the 2014 Conference on Empirical Methods in
  Natural Language Processing (EMNLP)}.

\bibitem[\protect\citename{Kingma and Ba}2014]{kingma2014adam}
Diederik Kingma and Jimmy Ba.
\newblock 2014.
\newblock Adam: A method for stochastic optimization.
\newblock {\em arXiv preprint arXiv:1412.6980}.

\bibitem[\protect\citename{Lazaridou \bgroup et al.\egroup
  }2013]{lazaridou-EtAl:2013:ACL2013}
Angeliki Lazaridou, Marco Marelli, Roberto Zamparelli, and Marco Baroni.
\newblock 2013.
\newblock Compositional-ly derived representations of morphologically complex
  words in distributional semantics.
\newblock In {\em Proceedings of the 51st Annual Meeting of the Association for
  Computational Linguistics (Volume 1: Long Papers)}, pages 1517--1526, Sofia,
  Bulgaria, August. Association for Computational Linguistics.

\bibitem[\protect\citename{Ling \bgroup et al.\egroup
  }2015a]{ling-EtAl:2015:EMNLP2}
Wang Ling, Chris Dyer, Alan~W Black, Isabel Trancoso, Ramon Fermandez, Silvio
  Amir, Luis Marujo, and Tiago Luis.
\newblock 2015a.
\newblock Finding function in form: Compositional character models for open
  vocabulary word representation.
\newblock In {\em Proceedings of the 2015 Conference on Empirical Methods in
  Natural Language Processing}, pages 1520--1530, Lisbon, Portugal, September.
  Association for Computational Linguistics.

\bibitem[\protect\citename{Ling \bgroup et al.\egroup }2015b]{ling2015charnmt}
Wang Ling, Isabel Trancoso, Chris Dyer, and Alan~W. Black.
\newblock 2015b.
\newblock Character-based neural machine translation.
\newblock {\em arXiv preprint arXiv:1511.04586}.

\bibitem[\protect\citename{Luong and Manning}2016]{luong2016}
Minh-Thang Luong and Christopher~D. Manning.
\newblock 2016.
\newblock Achieving open vocabulary neural machine translation with hybrid
  word-character models.
\newblock {\em arXiv preprint arXiv:1604.00788}.

\bibitem[\protect\citename{Luong \bgroup et al.\egroup
  }2013]{luong-socher-manning:2013:CoNLL-2013}
Thang Luong, Richard Socher, and Christopher Manning.
\newblock 2013.
\newblock Better word representations with recursive neural networks for
  morphology.
\newblock In {\em Proceedings of the Seventeenth Conference on Computational
  Natural Language Learning}, pages 104--113, Sofia, Bulgaria, August.
  Association for Computational Linguistics.

\bibitem[\protect\citename{Marelli \bgroup et al.\egroup
  }2014]{marelli2014semeval}
Marco Marelli, Luisa Bentivogli, Marco Baroni, Raffaella Bernardi, Stefano
  Menini, and Roberto Zamparelli.
\newblock 2014.
\newblock {SemEval}-2014 task 1: Evaluation of compositional distributional
  semantic models on full sentences through semantic relatedness and textual
  entailment.
\newblock In {\em Proceedings of the 8th International Workshop on Semantic
  Evaluation ({SemEval} 2014)}.

\bibitem[\protect\citename{Mitchell and Lapata}2010]{Mitchell:Lapata:2010}
Jeff Mitchell and Mirella Lapata.
\newblock 2010.
\newblock Composition in distributional models of semantics.
\newblock {\em Cognitive Science}, 34(8).

\bibitem[\protect\citename{Mrk{\v{s}}i{\'c} \bgroup et al.\egroup
  }2016]{mrkvsic2016counter}
Nikola Mrk{\v{s}}i{\'c}, Diarmuid~{\'O} S{\'e}aghdha, Blaise Thomson, Milica
  Ga{\v{s}}i{\'c}, Lina Rojas-Barahona, Pei-Hao Su, David Vandyke, Tsung-Hsien
  Wen, and Steve Young.
\newblock 2016.
\newblock Counter-fitting word vectors to linguistic constraints.
\newblock {\em arXiv preprint arXiv:1603.00892}.

\bibitem[\protect\citename{Pavlick \bgroup et al.\egroup
  }2015]{PavlickEtAl-2015:ACL:PPDB2.0}
Ellie Pavlick, Pushpendre Rastogi, Juri Ganitkevich, Benjamin~Van Durme, and
  Chris Callison-Burch.
\newblock 2015.
\newblock {PPDB} 2.0: Better paraphrase ranking, fine-grained entailment
  relations, word embeddings, and style classification.
\newblock In {\em Proceedings of the Annual Meeting of the Association for
  Computational Linguistics}.

\bibitem[\protect\citename{Pennington \bgroup et al.\egroup
  }2014]{pennington2014glove}
Jeffrey Pennington, Richard Socher, and Christopher~D Manning.
\newblock 2014.
\newblock Glove: Global vectors for word representation.
\newblock {\em Proceedings of Empirical Methods in Natural Language Processing
  (EMNLP 2014)}.

\bibitem[\protect\citename{Qiu \bgroup et al.\egroup
  }2014]{qiu-EtAl:2014:Coling1}
Siyu Qiu, Qing Cui, Jiang Bian, Bin Gao, and Tie-Yan Liu.
\newblock 2014.
\newblock Co-learning of word representations and morpheme representations.
\newblock In {\em Proceedings of COLING 2014, the 25th International Conference
  on Computational Linguistics: Technical Papers}, pages 141--150, Dublin,
  Ireland, August. Dublin City University and Association for Computational
  Linguistics.

\bibitem[\protect\citename{Schwartz \bgroup et al.\egroup
  }2015]{schwartz-reichart-rappoport:2015:Conll}
Roy Schwartz, Roi Reichart, and Ari Rappoport.
\newblock 2015.
\newblock Symmetric pattern based word embeddings for improved word similarity
  prediction.
\newblock In {\em Proceedings of CoNLL 2015}.

\bibitem[\protect\citename{Socher \bgroup et al.\egroup
  }2011]{SocherEtAl2011:PoolRAE}
Richard Socher, Eric~H Huang, Jeffrey Pennin, Christopher~D Manning, and
  Andrew~Y Ng.
\newblock 2011.
\newblock Dynamic pooling and unfolding recursive autoencoders for paraphrase
  detection.
\newblock In {\em Advances in Neural Information Processing Systems}.

\bibitem[\protect\citename{Socher \bgroup et al.\egroup }2013]{socher-13}
Richard Socher, Alex Perelygin, Jean Wu, Jason Chuang, Christopher~D. Manning,
  Andrew Ng, and Christopher Potts.
\newblock 2013.
\newblock Recursive deep models for semantic compositionality over a sentiment
  treebank.
\newblock In {\em Proceedings of the 2013 Conference on Empirical Methods in
  Natural Language Processing}.

\bibitem[\protect\citename{Soricut and Och}2015]{soricut2015unsupervised}
Radu Soricut and Franz Och.
\newblock 2015.
\newblock Unsupervised morphology induction using word embeddings.
\newblock In {\em Proc. NAACL}.

\bibitem[\protect\citename{Sperr \bgroup et al.\egroup
  }2013]{sperr-niehues-waibel:2013:CVSC}
Henning Sperr, Jan Niehues, and Alex Waibel.
\newblock 2013.
\newblock Letter n-gram-based input encoding for continuous space language
  models.
\newblock In {\em Proceedings of the Workshop on Continuous Vector Space Models
  and their Compositionality}, pages 30--39, Sofia, Bulgaria, August.
  Association for Computational Linguistics.

\bibitem[\protect\citename{Srivastava \bgroup et al.\egroup
  }2014]{srivastava2014dropout}
Nitish Srivastava, Geoffrey Hinton, Alex Krizhevsky, Ilya Sutskever, and Ruslan
  Salakhutdinov.
\newblock 2014.
\newblock Dropout: A simple way to prevent neural networks from overfitting.
\newblock {\em The Journal of Machine Learning Research}, 15(1):1929--1958.

\bibitem[\protect\citename{Sutskever \bgroup et al.\egroup
  }2011]{sutskever2011generating}
Ilya Sutskever, James Martens, and Geoffrey~E Hinton.
\newblock 2011.
\newblock Generating text with recurrent neural networks.
\newblock In {\em Proceedings of the 28th International Conference on Machine
  Learning (ICML-11)}, pages 1017--1024.

\bibitem[\protect\citename{Tai \bgroup et al.\egroup }2015]{tai2015improved}
Kai~Sheng Tai, Richard Socher, and Christopher~D Manning.
\newblock 2015.
\newblock Improved semantic representations from tree-structured long
  short-term memory networks.
\newblock {\em arXiv preprint arXiv:1503.00075}.

\bibitem[\protect\citename{{Theano Development
  Team}}2016]{2016arXiv160502688short}
{Theano Development Team}.
\newblock 2016.
\newblock {Theano: A {Python} framework for fast computation of mathematical
  expressions}.
\newblock {\em arXiv e-prints}, abs/1605.02688, May.

\bibitem[\protect\citename{Wieting \bgroup et al.\egroup
  }2015]{wieting2015ppdb-short}
John Wieting, Mohit Bansal, Kevin Gimpel, Karen Livescu, and Dan Roth.
\newblock 2015.
\newblock From paraphrase database to compositional paraphrase model and back.
\newblock {\em Transactions of the ACL (TACL)}.

\bibitem[\protect\citename{Wieting \bgroup et al.\egroup
  }2016]{wieting-16-full}
John Wieting, Mohit Bansal, Kevin Gimpel, and Karen Livescu.
\newblock 2016.
\newblock Towards universal paraphrastic sentence embeddings.
\newblock In {\em Proceedings of International Conference on Learning
  Representations}.

\bibitem[\protect\citename{Xu \bgroup et al.\egroup }2015]{xu2015semeval}
Wei Xu, Chris Callison-Burch, and William~B Dolan.
\newblock 2015.
\newblock {SemEval}-2015 task 1: Paraphrase and semantic similarity in
  {Twitter} ({PIT}).
\newblock In {\em Proceedings of the 9th International Workshop on Semantic
  Evaluation ({SemEval})}.

\bibitem[\protect\citename{Zhang \bgroup et al.\egroup
  }2015]{zhang-charcnn-2015}
Xiang Zhang, Junbo Zhao, and Yann LeCun.
\newblock 2015.
\newblock Character-level convolutional networks for text classification.
\newblock In {\em Advances in Neural Information Processing Systems 28}.

\end{thebibliography}
\bibliographystyle{emnlp2016}

\end{document}